*Article*

# Augmented Reality for Large Scene Based on a Unmarked Registration Framework

**Zhen Ma, He Xu \*, Yonghui Zhang, Junlong Chen, Dongbo Zhao, Siqing Chen**

College of Mechanical and Electrical Engineering, Harbin Engineering University, Harbin 150001, China;

\* Correspondence: railway_dragon@163.com; Tel:86-13351117608

**Abstract:** In this paper, a mobile camera positioning method based on kinematics of robot is proposed, which can realize far point positioning of imaging position and attitude tracking in large scene enhancement. Orbit precision motion through the framework overhead cameras and combining with the ground system of sensor array object such as mobile robot platform of various sensors, realize the good 3 d image registration, solve any artifacts that is mobile robot in the large space position initialization problem, effectively implement the large space no marks augmented reality(AR), human-computer interaction, and information summary. Finally, the feasibility and effectiveness of the method are verified by experiments.

**Keywords:** large-scale scene; augmented reality; unmarked; kinematics

## 1. Introduction

Augmented reality (AR) provides new tools for human-based control and interaction with complex systems. Users' sensory physical experience is combined with information from digital systems to achieve a good experience for users [1-4].In the process of industrial production, especially the assembly of large mechanical equipment such as aircraft, problems can be effectively solved through the AR of large scenes [5-7].In the AR display of large scenes, the marked method is very sensitive to factors such as light and visibility [8-13], e.g. under the condition of 220 lx-ray intensity, the maximum recognition distance of 80×80mm custom square sign is 1.70m, while the rectangular QR mark with a size of 80×400mm has a maximum recognition distance of only 1.58m under 220lx-ray intensity, so they are all not conducive to AR observation in complex environments [14-17].In the case of large scene blocks, only the stable tracking of the camera can accurately determine the superposition position of the AR and obtain a good visual experience [17-19].Problems and challenges are aimed, under the condition of large scene, and a kinematics method of AR is investigated, using positioning mobile camera and the sensor system, with kinematic transform method of the frame camera position.

Two demoes are presented to validated the AR approach in large scene by water hydraulic flexible manipulator and the mobile robot experiment for long-distance increase origin detailed scene of tracking and positioning, and in a large and touch screen to achieve the real-time information feedback of artifacts in the scene AR and interaction. The method is dramatically suitable for industrial environments with poor optical environment and interference conditions.

The paper is organized as follows. In section 2, the work of untagged AR will be covered.

In section 3, it describes the whole system, the theoretical model of coordinate transformation, the mathematical formula of virtual image position conversion and the generation of the screen image. In section 4, the hardware (mechanical, control) and software running process of the system are described. The section 5, two experiments are investigated to validate theoretical method, and the conclusions and future work are contained in section 6.

**2. Related Work**

SLAM is a typically untagged approach that tracks camera movements in real-time and builds virtual images in unprepared environments. Klein et al. [20] proposed breakthrough method by lens and real-time tracking system with large cumulative error and it is only suitable for scene tracking with small drift accumulation. Also, all these vision-based SLAM methods are still prone to system crashes due to moving blurred or text-based images [21-25].

To overcome the above single-lens vision obstacle [26], Wei fang et al. developed a real-time six-degree-of-freedom tracking method. A wide angle monocular camera and an inertial sensor are combined to achieve better robust six-degree-of-freedom motion tracking through mutual compensation between heterogeneous sensors. However, the method has a large cumulative drift error and weak anti-interference ability [27-33].

**3. Framework of AR with Unmarked Registration Based on Robotic Kinematics**

Based on the kinematics theory, the position points corresponding to the motion camera are obtained by the end position and attitude of the AR observation. High precision displacement sensor is used instead of the traditional visual recognition method with independency of good light, strong anti-interference ability and small cumulative error. It can be widely used in various large scenes such as the industrial environment.

A large scene AR method based on encoder track and sensors is presented. Different from the traditional SLAM, this scheme only requires the use of an encoded camera and sensors, and it can effectively solve the drift of the camera motion position and the accumulated error of the system in large scene AR scenarios.

*3.1. Kinematic Scene Coordinate Transformation Model*

To let the AR of virtual appear on the appropriate display position, a calculation method based on the kinematics coordinate system transformation on large space coordinate measuring system with unmarked modeling is proposed , and also the workpiece coordinate system with corresponding characteristic of camera image plane mapping between the model is also built with the gain and camera artifacts stance. These models will be imported into the unity 3D.

3.1.1 The Components of the System's Coordinate System

It is necessary to establish five coordinate systems as fellows.

1) The global coordinate system $C_w(O_w, X_w, Y_w, Z_w)$: attached on the orbital frame as the reference frame, it is used to describe the position and attitude of the camera, the position and attitude of the workpiece and that of the display screen. Points in it are represented as $P_W = (X_W, Y_W, Z_W)^T$.

2) The workpiece coordinate system $C_p(O_p, X_p, Y_p, Z_p)$: the origin $O_p$ of the frame is established on the first feature point of the workpiece, and the connection direction of the 1st and $P_5$ point is taken as the positive direction of axis Z; The axis Y is defined by the plane formed by points $P_1, P_3, P_5$ and the direction of axis X is determined by the right-hand rule. The points in it are represented as $P_p = (X_p, Y_p, Z_p)^T$.

3) Camera coordinate system $C_c(O_c, X_c, Y_c, Z_c)$: the origin $O_c$ is established at the perspective projection center of the camera with the axes X and Y correspond to the $X_c$ axis and $Y_c$-axis parallel to the image plane coordinate system. The points in it are noted as $P_c = (X_c, Y_c, Z_c)^T$.

4) Screen coordinate system: the origin $O_o$ is located at the upper left corner of the image plane. $(u, v)$ represents the image frame, while $C_o(O_o uv)$ with pixels as the unit. (x, y) represents the image frame $C_1(O_1 xy)$, whose origin $O_1$ is located at the intersection of the camera optical axis and the image plane. The X-axis and Y-axis are parallel to the horizontal and vertical axes of the pixel respectively. $O_1$ in the $C_0$ coordinates is $(u_0, v_0)$.

5) Camera cradle frame coordinate system $C_r(O_r, X_r, Y_r, Z_r)$: the origin $O_r$ is established at the intersection of the two rotating axes of the camera turntable, and the rotation angles of $Z_r$ and $Y_r$ of the two rotating axes at a certain time are used to determine the attitude of the camera turntable coordinate system $C_r$ in the reference system $C_W$. There is a rigid connection between the camera and the camera turntable.

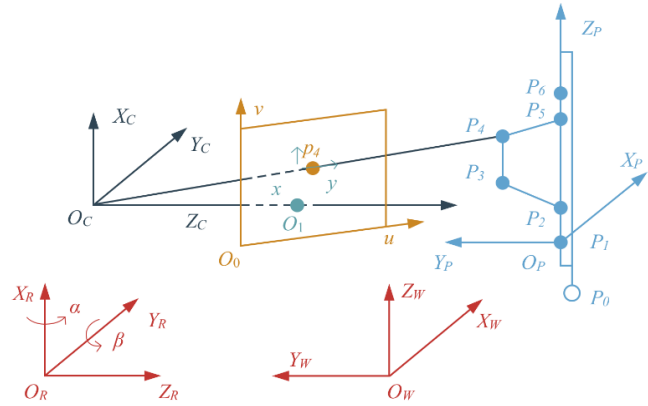

**Figure 1.** The model of coordinate system

The camera coordinate system and the cradle head coordinate system are fixed and constant. The spatial points in the coordinate system of the turntable are represented as $P_r = (X_r, Y_r, Z_r)^T$.

3.1.2 Transformation of Coordinate Points

Assuming that the coordinates of the space point in the two coordinate systems are respectively $P_1 = (X_1, Y_1, Z_1)^T$ and $P_2 = (X_2, Y_2, Z_2)^T$, then the following formula for the pose transformation can be obtained:

$$P_1 = RP_2 + T \tag{1}$$

where $R = \begin{bmatrix} r_{11} & r_{12} & r_{12} \\ r_{21} & r_{22} & r_{23} \\ r_{31} & r_{32} & r_{33} \end{bmatrix}$ and $T = (t_x, t_y, t_z)^T$.

The point is represented by the homogeneous coordinate $P = (X, Y, Z, 1)^T$, and the pose transformation formula is converted into the following form:

$$P_1 = \begin{bmatrix} R & T \\ 0 & 1 \end{bmatrix} P_2 = M_{21} P_2 \tag{2}$$

where $M_{12}$ is the pose matrix from frame 2 to frame 1.

In figure 2, Euler angles are noted by three independent angles (row, position, and angle).

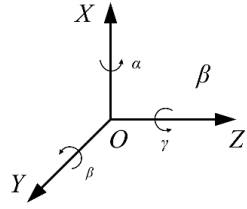

**Figure 2.** Euler angle parameterization

Under certain conditions, any attitude of the a frame system can be uniquely represented by a set of Euler angles:

$$R_{xy} = \begin{bmatrix} \cos\beta\cos\gamma & \sin\alpha\sin\beta\cos\gamma - \cos\alpha\sin\gamma & \cos\alpha\sin\beta\cos\gamma + \sin\alpha\sin\gamma \\ \cos\beta\sin\gamma & \sin\alpha\sin\beta\sin\gamma + \cos\alpha\cos\gamma & \cos\alpha\sin\beta\sin\gamma - \sin\alpha\cos\gamma \\ -\sin\beta & \sin\alpha\cos\beta & \cos\alpha\cos\beta \end{bmatrix} \tag{3}$$

All the coordinate calculations will be completed in unity, and the relational variables will be determined by the initialization measurement. In the process, as long as coordinate position and the displacement vector are imported, the world coordinate system corresponding to the points of other coordinates will calculated.

**3.2. Generation and Processing of Screen Images**

The camera model comes from the transform of the point in global coordinate system into the those in image plane coordinate system. The transformation formula $P_W = (X_W, Y_W, Z_W)^T$ to image coordinate system $C_0 = (u, v)^T$ in the global coordinate system is:

$$C_0 = PP_w \tag{4}$$

where $P = M_1 [R | T]$, $M_1$ is perspective projection matrix expressed by internal and external parameter matrices. It includes the center point (ideally the center of the image), the actual focal length, lens distortion (mainly include radial distortion and tangential distortion) and other systematic error parameters, and the form of the internal parameter matrix is as :

$$M_1 = \begin{bmatrix} f_x & R & u_0 \\ 0 & f_y & v_0 \\ 0 & 0 & 1 \end{bmatrix} \quad (5)$$

where, $f_x$ and $f_y$ are focal lengths in the x and y directions, respectively. $R$ is the distortion factor. $u_0$ and $v_0$ are the center coordinates of the graph.

$R_{xy}$ is the rotation change matrix and T is the translation matrix.

$$[R_{xy} | T] = \begin{bmatrix} \cos\beta\cos\gamma & \sin\alpha\sin\beta\cos\gamma - \cos\alpha\sin\gamma & \cos\alpha\sin\beta\cos\gamma + \sin\alpha\sin\gamma | t_x \\ \cos\beta\sin\gamma & \sin\alpha\sin\beta\sin\gamma + \cos\alpha\cos\gamma & \cos\alpha\sin\beta\sin\gamma - \sin\alpha\cos\gamma | t_y \\ -\sin\beta & \sin\alpha\cos\beta & \cos\alpha\cos\beta & | t_z \end{bmatrix} \quad (6)$$

The relationship between a point in 3D space and its projected pixel points on the image space can be described in detail as :

$$\begin{bmatrix} x_i \\ y_i \\ 1 \end{bmatrix} = \begin{bmatrix} f_x & r & u_0 \\ 0 & f_y & v_0 \\ 0 & 0 & 1 \end{bmatrix} \begin{bmatrix} \cos\beta\cos\gamma & \sin\alpha\sin\beta\cos\gamma - \cos\alpha\sin\gamma & \cos\alpha\sin\beta\cos\gamma + \sin\alpha\sin\gamma | t_x \\ \cos\beta\sin\gamma & \sin\alpha\sin\beta\sin\gamma + \cos\alpha\cos\gamma & \cos\alpha\sin\beta\sin\gamma - \sin\alpha\cos\gamma | t_y \\ -\sin\beta & \sin\alpha\cos\beta & \cos\alpha\cos\beta & | t_z \end{bmatrix} \begin{bmatrix} X_i \\ Y_i \\ Z_i \\ 1 \end{bmatrix} \quad (7)$$

The internal parameter matrix of the camera is the inherent parameter of the camera which can be obtained by calibrating. Although the transformation matrix has 12 unknowns, it has only 11 degrees of freedom. Only 6 pairs of points in 3D space and their corresponding 2D image points are needed to solve the 12 parameters. These parameters include the camera's position and attitude. According to matrix multiplication, the following forms can be obtained:

$$\begin{bmatrix} x_i \\ y_i \\ 1 \end{bmatrix} = \begin{bmatrix} \cos\beta\cos\gamma X_i + \sin\alpha\sin\beta\cos\gamma Y_i - \cos\alpha\sin\gamma Y_i + \cos\alpha\sin\beta\cos\gamma Z_i + \sin\alpha\sin\gamma Z_i + t_x \\ \cos\beta\sin\gamma X_i + \sin\alpha\sin\beta\sin\gamma Y_i + \cos\alpha\cos\gamma Y_i + \cos\alpha\sin\beta\sin\gamma Z_i - \sin\alpha\cos\gamma Z_i + t_x \\ -\sin\beta X_i + \sin\alpha\cos\beta Y_i + \cos\alpha\cos\beta Z_i + t_z \end{bmatrix} \quad (8)$$

The coordinate value of the target on the image is:

$$\begin{cases} x_i = \dfrac{\cos\beta\cos\gamma X_i + \sin\alpha\sin\beta\cos\gamma Y_i - \cos\alpha\sin\gamma Y_i + \cos\alpha\sin\beta\cos\gamma Z_i + \sin\alpha\sin\gamma Z_i + t_x}{-\sin\beta X_i + \sin\alpha\cos\beta Y_i + \cos\alpha\cos\beta Z_i + t_z} \\ y_i = \dfrac{\cos\beta\sin\gamma X_i + \sin\alpha\sin\beta\sin\gamma Y_i + \cos\alpha\cos\gamma Y_i + \cos\alpha\sin\beta\sin\gamma Z_i - \sin\alpha\cos\gamma Z_i + t_x}{-\sin\beta X_i + \sin\alpha\cos\beta Y_i + \cos\alpha\cos\beta Z_i + t_z} \end{cases} \quad (9)$$

Homogeneous linear equations can be solved by removing the denominator, and two such equations can be obtained for each pair of points. Finding the position of $N$ different points in the world coordinate system and determining their coordinates in the camera image, homogeneous linear equations can be solved, and two such equations can be obtained for each pair of points. When $N \geq 6$, the position and attitude of the AR virtual image embedded in the whole system can be solved.

**4. Design and Composition of the System**

*4.1 Design of Mechanical Structure*

Figure 3 shows an orbiting mobile camera designed by the experimental team at Harbin engineering university. Mobile camera adopts the method of bilateral orbit and upper and lower clamping balance, camera movement is rolling on the sprocket in the chain, sprocket drive is done by stepping motor, encoder read the rotation of the sprocket wheel and then calculate the movement of the camera position, on a host computer through wireless signal to control the operation of the whole system. The trolley mechanical system consists of five parts: frame and chain mechanism, sliding-contact line power supply mechanism, sprocket drive mechanism and encoder feedback mechanism.

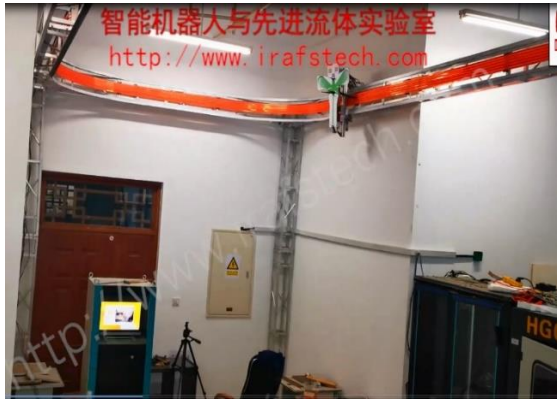 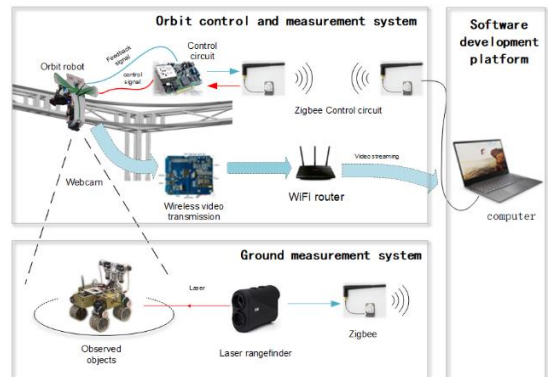

**Figure 3.** The structure of the mobile camera system       **Figure 4.** The control system architecture.

*4.2 Control System Design*

The control system has three functions :(a) according to the input signal of the computer to control the position, speed, switch signal and other information of the motion camera; (b) receive and display the video signal returned by the camera; C) receive and calculate the data of the rangefinder. As shown in figure 4.

The main component of the controller is STM32 single-chip microcomputer. On the one hand, it receives the control signal from the computer and sends out the pulse signal to drive the stepper motor. The other receives feedback from the encoder and sends it to the computer for processing. To realize barrier-free operation in the air, it communicates with the computer through Zigbee wireless transmission and uses sliding contact line for sliding power supply.

*4.3 Software Generation of The System*

Unity3D software is used as the development platform of AR in this paper. On the one hand, Unity3D can obtain the video stream by executing the C# script and display the images that can reflect the real world on the screen. On the other hand, the camera composition information from the sensor is also received by the executing code, and the attitude of the virtual object in the display screen is expressed through the corresponding algorithm (described in section 4). Therefore, virtual objects are presented on the screen for AR. The design flow of the software is determined as shown in figure 5.

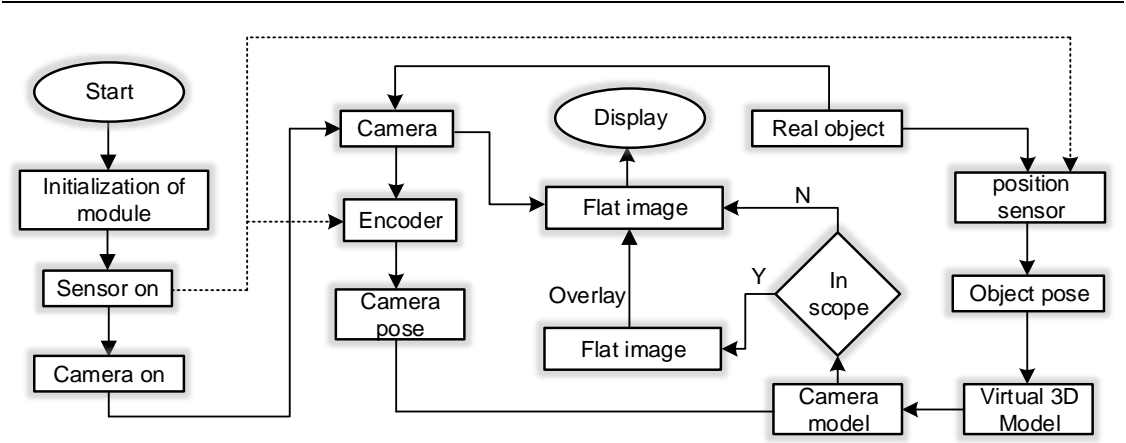

**Figure 5.** Flowchart of our proposed method.

## 5. Error Analysis and Experiment

The experimental equipment is shown in figure 6 and the software running process of the system is shown in figure 7.

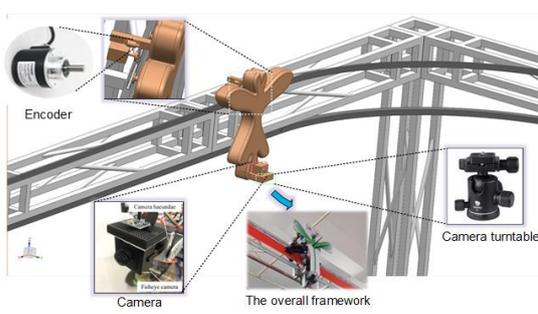
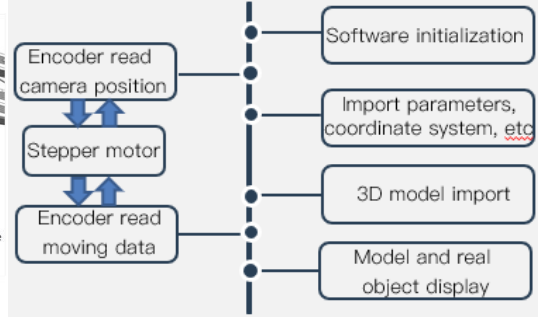

**Figure 6.** Equipment structure diagram.   **Figure 7.** Software running process of the system.

In this experiment, the camera track is a runway, 4 meters high from the ground, 6 meters long and 3 meters wide, with four turning radii of 1 meter. Z880A industrial camera and sy-03 manual turntable is used. The camera uses a fixed focal length with the resolution and frame rate set to 640*480:30 frames. Bes38-0656n-1000 encoder is adopted.

### 5.1. Analysis of The Error of the Equipment System

To independently calculate the influence of system structure error on AR observation, coordinate measurement error and calibration error are not considered in this experimental analysis. Therefore, the coordinate transformation relation is as follows:

$$\begin{bmatrix} X_{W0} \\ Y_{W0} \\ Z_{W0} \\ 1 \end{bmatrix} = \begin{bmatrix} \cos\beta & \sin\alpha\sin\beta & \cos\alpha\sin\beta & t_x \\ 0 & \cos\alpha & -\sin\alpha & t_y \\ -\sin\beta & \sin\alpha\cos\beta & \cos\alpha\cos\beta & t_z \\ 0 & 0 & 0 & 1 \end{bmatrix} M_{cr} \begin{bmatrix} X_{c0} \\ Y_{c0} \\ Z_{c0} \\ 1 \end{bmatrix} \quad (10)$$

Assuming $M_{cr} = 1$, then

$$\begin{cases} X_{w0} = \cos\beta X_{c0} + \sin\alpha\sin\beta Y_{c0} + \cos\alpha\sin\beta Z_{c0} + t_x \\ Y_{w0} = \cos\alpha Y_{c0} - \sin\alpha Z_{c0} + t_y \\ Z_{w0} = -\sin\beta X_{c0} + \sin\alpha\cos\beta Y_{c0} + \cos\alpha\cos\beta Z_{c0} + t_z \end{cases} \quad (11)$$

The error transfer coefficient is

$$\begin{cases} \dfrac{\partial X_{w0}}{\partial \alpha} = \cos\alpha\sin\beta Y_{c0} - \sin\alpha\sin\beta Z_{c0}, \\ \dfrac{\partial X_{w0}}{\partial \beta} = \sin\beta X_{c0} - \sin\alpha\cos\beta Y_{c0} - \cos\alpha\cos\beta Z_{c0} \\ \dfrac{\partial Y_{w0}}{\partial \alpha} = \sin\alpha Y_{c0} + \cos\alpha Z_{c0}, \dfrac{\partial Y_{w0}}{\partial \beta} = 0, \dfrac{\partial Z_{w0}}{\partial t_z} = 1 \\ \dfrac{\partial Y_{w0}}{\partial t_y} = 1, \dfrac{\partial Z_{w0}}{\partial \alpha} = -\cos\alpha\cos\beta Y_{c0} - \sin\alpha\cos\beta Z_{c0} \\ \dfrac{\partial Z_{w0}}{\partial \beta} = -\cos\beta X_{c0} + \sin\alpha\sin\beta Y_{c0} + \cos\alpha\sin\beta Z_{c0} \end{cases} \quad (12)$$

Assuming that the standard deviation of the camera turntable angle and the orbital translation output are independent to each other, the standard deviation of the coordinate measurement can be obtained according to the error transfer relation.

$$\begin{cases} \sigma_{X_{w0}} = \sqrt{(\cos\alpha\sin\beta Y_{c0} - \sin\alpha\sin\beta Z_{c0})^2 \sigma_\alpha^2 + (\sin\beta X_{c0} - \sin\alpha\cos\beta Y_{c0} - \cos\alpha\cos\beta Z_{c0})^2 \sigma_\beta^2 + \sigma_{t_x}^2} \\ \sigma_{Y_{w0}} = \sqrt{(\sin\alpha Y_{c0} + \cos\alpha Z_{c0})^2 \sigma_\alpha^2 + \sigma_{t_y}^2} \\ \sigma_{z_{w0}} = \sqrt{(\cos\alpha\cos\beta Y_{c0} + \sin\alpha\cos\beta Z_{c0})^2 \sigma_\alpha^2 + (-\cos\beta X_{c0} + \sin\alpha\sin\beta Y_{c0} + \cos\alpha\sin\beta Z_{c0})^2 \sigma_\beta^2 + \sigma_{t_z}^2} \end{cases} \quad (13)$$

Analyzing the influence of the output error of the camera turntable's angle on the observation when the orbit moves and the camera turntable rotates to different angles. It can be seen that the system error is related to the accuracy of the orbit and the output error of the camera turntable angle, as well as the rotation angle of the camera turntable and the observation distance. In this experiment, the observation and distance measurement $Z$=3000mm, the Z direction of the orbit is parallel to the ground adjustment, the maximum angle of the experiment is 45°, the values of $X$ and $Y$ are 100mm, and the error transfer coefficient is set as $\sigma_{X_{w0}} = 1.51, \sigma_{Y_{w0}} = 1.55, \sigma_{z_{w0}} = 1.13$.

*5.2 Static Workpiece Experiment*

The experiment completed the test by making 30 repeated observations of the water hydraulic flexible manipulator. Three observation points were selected on the orbit of 0mm, 300mm, 1500mm, and repeatability tests were conducted on each point. Before the experiment, the calibration module in Unity3D software was used to calibrate the camera and cradle head, and the internal parameters of the camera were calibrated. The results are shown in Table 1.

**Table 1**. The camera parameters

| Camera Model | $f_x$ | $f_x$ | $r$ | Pixel | | Distortion Coefficient | |
|---|---|---|---|---|---|---|---|
| | | | | $u_0$ | $v_0$ | $K_{.1}$ | $K_{.2}$ |
| Z100 | 3676.462 | 3676.478 | 0.263 | 645.342 | 508.259 | 1.30 | 1.88 |

The camera turntable is a 6-degree-of-freedom manual camera turntable, and the positioning accuracy is repeated to the indexing level, but only two of them are used in this experiment. After the camera is installed on the camera turntable with calibrated posture, the result is as:

$$M_{cr} = \begin{bmatrix} 0.9907 & 0.1353 & -0.0064 & 50.843 \\ -0.1396 & 0.9915 & 0.0093 & 47.094 \\ 0.0083 & -0.0085 & 0.9990 & 76.177 \\ 0 & 0 & 0 & 1 \end{bmatrix} \quad (14)$$

As shown in figure 8, the virtual image and the solid image are displayed in a fixed parallel direction, with a distance between them in the X-axis direction. The moving speed of the motion camera is 0.1m/s. The workpiece frame is obtained by measuring the feature points on the workpiece with the rangefinder.

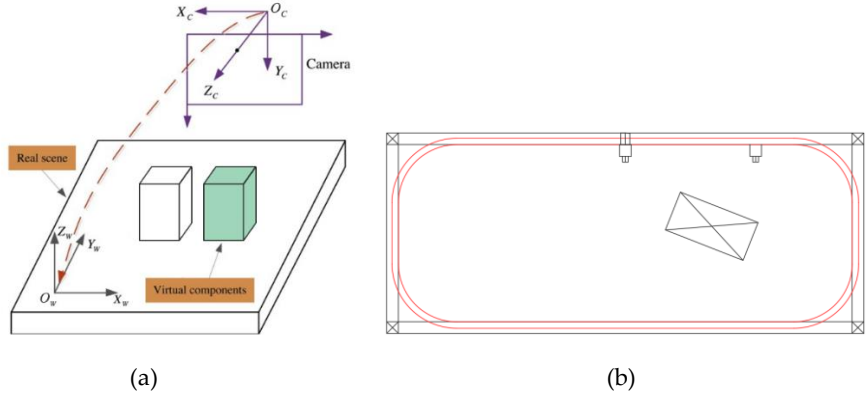

(a)            (b)

**Figure 8.** Schematic diagram of the first experiment. (**a**) Diagram for AR registration (**b**) Diagram camera movement

The initial state (zero point) of the workpiece is the coincidence with the virtual image and the physical image.

The first observation points: the virtual image to the moving direction of 300 mm x-axis, using video image measuring tool software and entity scale, measure the virtual image geometric center, center of plane geometry and the real image and actual distance on the x-axis.

The second observation point: move the camera to the moving direction of 2000 mm x-axis, virtual like settings, still, distance measuring geometric center. Initial state, first observation point, and second observation point are shown in figure 9.

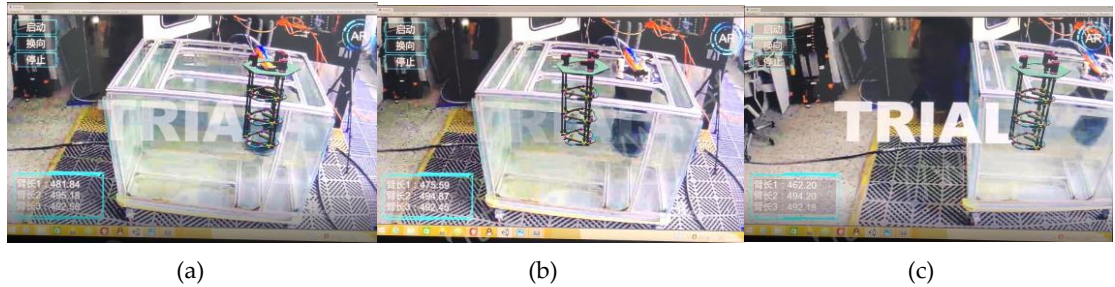

(a)          (b)          (c)

**Figure 9.** Screen display during the experiment. (**a**) Initialization display (**b**) First screenshot (**c**) Second screenshot

Table 2. Measuring result.

| The camera position | Camera range | Measurement of the mean | The standard deviation |
|---|---|---|---|
| Initialize the point | 0 mm | 0.5 mm | 1.20 |
| The first point | 300 mm | 288.9 mm | 1.35 |
| The second point | 2000 mm | 1499.5 mm | 1.41 |

It can be seen from the table that the moving distance of the moving camera has a certain influence on the observation. Because the manual camera turntable is used during small angle in this experiment, the error generated by the moving distance has a small influence. The observation accuracy can be significantly improved by increasing the orbit accuracy, minimizing the dynamic change of the turntable and decreasing the rotation angle.

*5.3 Moving workpiece experiment*

Using the same observation equipment as in the first experiment, the mobile camera was used to track the robot. The robot moves in a uniform straight line along the *X*-axis, and the moving camera tracks at the same speed to keep the distance between the virtual image and the solid image constant, as shown in figure 10.

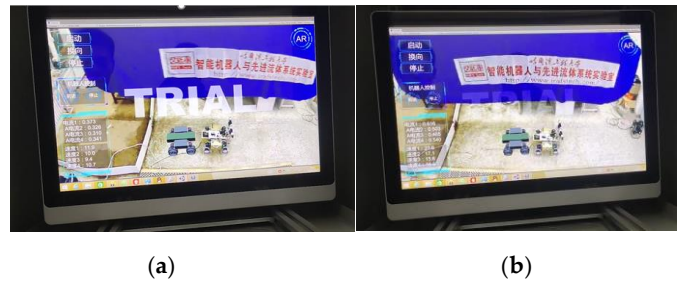

(**a**)          (**b**)

**Figure 10.** Experimental process display. (**a**) Start measuring screenshots (**b**) End measurement screenshot

The video image measurement tool software and solid ruler are still used to measure the change in the distance between the geometric center point of the virtual image and the solid image. The data result is shown in figure 11.

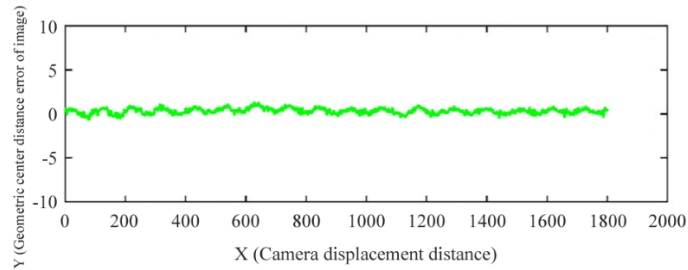

**Figure 11.** Error of real-time distance.

For comparison with a marked visual system, the standard deviation *R* is referenced.

$$R = \sqrt{\frac{1}{N}\sum_{i=1}^{N}(p_N - p_E)^2} \quad (15)$$

where, *N* is the number of samples under the total trajectory length, and the measured true error value is $P_N$, the theoretical error value $P_E$ =0 in this experiment. The actual walking length of the robot is more than 1500mm, so the first 1500mm will be taken as the data sample. Collect 30 points and calculate *R* = 0.5%, which is more accurate than visual tracking.

**6.Conclusions**

In this paper, an unmarked large Scene of view AR observation method is proposed, which effectively solves the problem of light sensitivity of labeled AR. As a new method, it is especially suitable for the industrial plant with strong anti-interference ability. Finally, two experiments are carried out, and the results show that the system can well enhance the display of reality images in large space, make the images always in the right position and size, and has good adaptability.

However, this paper focuses on the calculation principle of AR and spatial conversion brought by the encoder, as well as error analysis. The virtual image generated by the experiment is fixed relative to the entity, and will not change with time, camera movement and other factors. In the future, dynamic algorithms will be used to improve the performance of AR displays.

**Author Contributions**: Zhen Ma. conceived and designed the system and experiments; He Xu performed conceptualization and writing; Yonghui Zhang,Junlong Chen, Dongbo Zhao, Siqing Chen finished the communication softwawre,3D unity programming and control system respectively.

**Funding:** This work was partially supported by the Natural Science Foundation of China under Grant 51875113，Natural Science Joint Guidance Foundation of the Heilongjiang Province of China under Grant LH2019E027,"Jinshan Talent" Zhenjiang Manufacture 2025 Leading Talent Project,"Jiangyan Planning" Project in Yangzhong City.

**Conflicts of Interest:** The authors declare no conflict of interest.

**References**

1. Levi Manring, John Pederson, Dillon Potts, Beth Boardman, David Mascarenas, Troy Harden and Alessandro Cattaneo; Augmented Reality for Interactive Robot Control; Modern Information Technology. Special switchable viewer in Structural Dynamics and Experimental Techniques, Volume 5, Conference Proceedings of the Society for Experimental Mechanics Series, https://doi.org/10.1007/978-3-030-12243-0_2
2. LIN Ruizong; PENG Chuanxiang; GAO Xian; LI Zhenglin.; Research on the Completion Acceptance of Substation Engineering -based on AR Space Measurement Technology; Modern Information Technology 2096-4706 (2018) 12-0190-03 Author 1, A.; Author 2, B. *Book Title*, 3rd ed.; Publisher: Publisher Location, Country, 2008; pp. 154–196.
3. Mar ı a - Blanca Ibanez, Angela Di - Serio; Diego Villaran - Molina; Carlos Delgado - Kloos; Support for Augmented Reality Simulation Systems: The Effects of Scaffolding on Learning Outcomes and Behavior Patterns; IEEE TRANSACTIONS ON LEARNING TECHNOLOGIES, vol.9, NO. 1, january-march 2016.
4. Yang Wang; Yuji Sato; Mai Otsuki; Hideaki Kuzuoka,Yusuke Suzuki;Effect of Body Representation Level of an Avatar on Quality of ar-based Remote Instruction; Multimodal Technol. Interact. 2020, 4, 3;Doi: 10.3390 / mti4010003
5. Industry 4.0 ongoing: how augmented reality is transforming manufacturing. http://www.cnki.net
6. CHEN Junhua; CHEN, Changyu; SUN Gang; WAN Bile. Research on the Application Technology of Augmented Reality in Spacecraft Cable Assembly; Science and Technology & Innovation 2095-6835(2019)08-0040-05
7. LUO Kang; LI Xin; Implementation of 3D Display of Products Based on Augmented Reality Technology; JOURNAL OF JINLING INSTITUTE OF TECHNOLOGY 1672-755x (2019)01-0006-05
8. Changmin Lim; Chanran Kim; jong-l Park; Hanhoon Park; Mobile Augmented Reality Based on Invisible Marker; IEEE International Symposium on Mixed and Augmented Reality Adjunct Proceedings. 978-1-5,090-3740-7/16
9. Robert Woll; Thomas Damerau; Kevin Wrasse,; Rainer Stark; Augmented Reality in a Serious Game for Manual Assembly Processes; IEEE International Symposium on Mixed and Augmented Reality 2011Science and Technology Proceedings 26-29 October, Basel, Switzerland.978-1-4673-0059-9/10

10. John Sausman; Alexei Samoylov; Susan Harkness Regli; Meredith Hopps; Effect of Eye and Body Movement on Augmented Reality in the Manufacturing Domain; IEEE International Symposium on Mixed and Augmented Reality 2012Science and Technology Proceedings 5-8 November 2012, Atlanta, Georgia 978-1-4673-4662-7/12
11. DIMITRIS CHATZOPOULOS; CARLOS BERMEJO; ZHANPENG HUANG; PAN HUI; Mobile Augmented Reality Survey: From Where We Are to Where We Go; IEEE Access Digital Object Identifier/Access. 10.1109 2017.2698164
12. Shao Jiang; Yan ketong; Yao Jun; Niu yafeng; Experimental study on visual search for aiming symbols in head-mounted AR system interface; JOURNAL OF SOUTHEAST UNIVERSITY (Natural Science Edition) 1001-0505(2020)01-0020-06
13. Ivo Maly; David Sedla 'cek; Paulo Leitao; Augmented Reality Experiments with Industrial Robot in Industry 4.0 Environment; The 978-1-5-090-2870 $31.00/2/16 © 2016 IEEE
14. Holger Regenbrecht; Industrial Augmented Realities Applications; University of Otago, New Zealand Chapter XIV
15. PAULA FRAGA-LAMAS; MIGUEL A. VILAR-MONTESINOS; A Practical Evaluation of Commercial Industrial Augmented Reality Systems in an Industry 4.0 Shipyard; SPECIAL SECTION ON HUMAN-CENTERED SMART SYSTEMS AND TECHNOLOGIES
16. V. Havard; D. Baudry; A. Louis, and B. Mazari; Augmented reality maintenance demonstrator and associated modelling, in Proc. IEEE Virtual Reality (VR), Mar. 2015, pp. 329–330.
17. Virtual Reality Technology Transforms Design of UK Warships. Accessed: Dec. 1, 2017. [Online]. Available: http://www.baesystems.com/en/article/virtual-reality-technology-transforms-design-of-uk-warships
18. LIU Rui; YIN Xu ;FAN Xiu;WANG Lei.HE Qi chang; AR Guidance Based visual inspection method for cable laying quality consistency check; Computer Integrated Manufacturing Systems
19. Zhenliang Zhang; Dongdong Weng; Haiyan Jiang; Yue Liu; Yongtian Wang; Inverse Augmented Reality: A Virtual Agent's Perspective; IEEE International Symposium on Mixed and Augmented Reality Adjunct 2018 (ismar-adjunct) 987-1-5386-7592-2/18
20. Alex Hill; Harrison Leach; Interactive Panned and Zoomed Augmented Reality Interactions Using COTS Heads Up Displays; IEEE International Symposium on Mixed an Augmented Reality 2014 Science and Technology Proceedings 10-12 September 2014,Munich.Germany.978-1-4799-6184-9/13
21. Doree Duncan Seligmann; Steve Feiner; Automated generation of intent - -based 3 d Illustrations, ACM SIGGRAPH Computer Graphics, v. 25 n. 4, p. 123-132, out of 1991
22. Dirk Groten; Ronald Van Der Lingen; Layar On Google Glass: It 's Not Augmented Reality, Web Blog Posting, June 25, 2013, https://www.layar.com/news/blog/2013/06/25/layar-on-google-glass
23. Smart Glasses Moverio BT - 200, Epson, http://www.epson.com/cgibin/Store/jsp/Landing/moverio-bt-200-smart-glasses.do
24. Christopher Freeman; Rab Scott; An exercise in cost and waste reduction using Augmented Reality in Composite Layup manufacturing; March 23-27, 2015 IEEE Virtual Reality Conference, Arles, France 978-1-4799-1727-3/15
25. Husam A.A das; Sachin Shetty.S.K eith Hargrove; Virtual and Augmented Reality Based Assembly Design System for Personalized Learning. Science and Information Conference on October 7-9, 2013
26. Ma Jindun; Zhang Lei; Guo Libin; Zhang Jie; Research Summary of Augmented Reality Equipment Maintenance Guidance System; Journal of Ordnance Equipment Engineering; ISSN 2096-2034, 50-1213 / CN TJ
27. Yaxuan Zhou; Paul Yoo; Yingru Feng; Aditya Sankar, Alireza Sadr, Eric j. Seibel; Towards ar-assisted visualisation and guidance for imaging of dental decay; Healthcare Technology Letters, 2019, vol.6, Iss. 6, pp. 243 -- 248 doi: 10.1049/ htr.2019.0082
28. Mohammed A, Schmidt B, Wang l. Active collision avoidance for human-robot collaboration driven by vision sensors[J]. International Journal of Computer Integrated Manufacturing, 2017, 30(9):970-980.
29. Wei Fang; Lianyu Zheng; Xiangyong Wu; Multi-sensor based real-time 6-dof pose tracking for wearable wearable reality; Computers in Industry. 92-93 (2017) 91-103

30. Wang. A comprehensive survey of augmented reality assembly research [J]. Advances in Manufacturing, 2016:1-22.
31. LI Jin; LIU Xuan; ZHANG Jianhua; ZHANG Jie; CHEN Hao; ZHANG Yaonan; Research on Indirect Solution of Inverse Kinematics Based on RBF Neural Network; The MACHINE TOOL & HYDRAULICS; ISSN1001-3881.20199.23.007
32. G. Klein, d. Murray; Parallel tracking and mapping for small AR workspaces,Proc. Of IEEE/ACM International Symposium on Mixed and Augmented Reality(2007) 1-10.
33. F. Endres, j. Hess, n. Engelhard, j. Sturm, d. Cremers, w. Burgard, An evaluationof the rgg-d SLAM system, Proc of IEEE International Conference on Robotics and Automation (2012) 1691 -- 1696.